\documentclass[conference]{IEEEtran}
	\PassOptionsToPackage{bookmarks={false}}{hyperref}
		
	\usepackage{chngpage}	
	\usepackage{float}	
	\usepackage[hyphens]{url}
	\usepackage{ctable}
	\usepackage[flushleft]{threeparttable}
	\usepackage{comment}
	\usepackage[subject={Todo}]{pdfcomment}
	\usepackage{listings}
	\usepackage{xcolor} 
	\usepackage{algorithm2e}

	\usepackage[font=bf,skip=\baselineskip]{caption}

\ifCLASSINFOpdf
\else
\fi

\usepackage{listings}
\usepackage{graphicx}
\usepackage[hyphens]{url}

\lstdefinestyle{turtle}{%
    morekeywords={a, @prefix},
    morecomment=[s][\textrm]{<}{>},
    morecomment=[s][\textit]{"}{"},
}

\newcommand{\furl}[1]{\footnote{\scriptsize \url{#1}}}

\begin{document}
%
\title{Git4Voc: Git-based Versioning for Collaborative Vocabulary Development}


\author{\IEEEauthorblockN{Lavdim Halilaj}
\IEEEauthorblockA{EIS Department\\
University of Bonn\\
Bonn, Germany\\
halilaj@cs.uni-bonn.de}
\and
\IEEEauthorblockN{Irl\'an Grangel-Gonz\'alez}
\IEEEauthorblockA{EIS Department\\
University of Bonn\\
Bonn, Germany\\
grangel@cs.uni-bonn.de}
\and
\IEEEauthorblockN{G{\"o}khan Coskun}
\IEEEauthorblockA{EIS Department\\
University of Bonn\\
Bonn, Germany\\
coskun@cs.uni-bonn.de}
\and
\IEEEauthorblockN{S{\"o}ren Auer}
\IEEEauthorblockA{EIS Department\\
University of Bonn\\
Bonn, Germany\\
auer@cs.uni-bonn.de}
}

\maketitle

\begin{abstract}
Collaborative vocabulary development in the context of data integration is the process of finding consensus between the experts of the different systems and domains. 
The complexity of this process is increased with the number of involved people, the variety of the systems to be integrated and the dynamics of their domain.
In this paper we advocate that the realization of a powerful version control system is the heart of the problem.
Driven by this idea and the success of Git in the context of software development, we investigate the applicability of Git for collaborative vocabulary development. 
Even though vocabulary development and software development have much more similarities than differences there are still important differences.
These need to be considered within the development of a successful versioning and collaboration system for vocabulary development.
Therefore, this paper starts by presenting the challenges we were faced with during the creation of vocabularies collaboratively and discusses its distinction to software development.
Based on these insights we propose Git4Voc which comprises guidelines how Git can be adopted to vocabulary development.
Finally, we demonstrate how Git hooks can be implemented to go beyond the plain functionality of Git by realizing vocabulary-specific features like syntactic validation and semantic diffs.
\end{abstract}


\section{Introduction}

One of the key obstacles for the wider deployment of semantic technologies is the lack of comprehensive vocabularies.
This is because vocabulary development requires a significant investment, which is difficult to make by a single person or organisation. 
If we look at current vocabularies (e.g. LOV\footnote{\url{http://lov.okfn.org}}), we observe that they are rather simplistic. 
For a total of 457 vocabularies listed in LOV, a straightforward query against the LOV SPARQL endpoint tells us that the average number of classes for each vocabulary is 42 whereas the average number of properties is 59. 
Omitting the four vocabularies with the highest number of classes and properties, these figures decrease to 31 classes and 37 properties on average. 
We also observe that a large number of crucial domains is not or only superficially covered by existing vocabularies. 
One of the main reasons for the lack of vocabularies is also the lack of adequate methodological and tool support.

At the same time, the problem of integrating data from different systems receives ever-increasing attention. 
Identifying the main terms across heterogeneous data sources by finding a consensus between the developers and defining a shared vocabulary is an effective approach to tackle this problem.
However, this process, which we refer to as \emph{collaborative vocabulary development}, itself is a complex problem to be solved. 
In fact, the main challenge for the vocabulary engineers is to work collaboratively on a shared objective in a harmonic and efficient way while avoiding misunderstandings, uncertainty and ambiguity.
The quality of the produced vocabularies is another challenge that should be tackled as well.
In~\cite{grangel2015Convention}, we identified and elaborated important aspects for vocabulary development such as: reuse, vocabulary structure, naming conventions, multilinguality, documentation, validation and authoring. 
These aspects are relevant from collaborative point of view as well.
Taking into consideration above aspects will impact the quality of vocabulary itself.

Therefore, finding a suitable collaboration methodology is exacerbated by the number and diversity of the involved stakeholders as well as the complexity of the domains.
Due to the open, distributed and participatory nature of the Web, such a solution is of paramount interest for the Semantic Web community.

Our approach to tackle the mentioned problem is to focus on supporting the collaborative vocabulary development with a well-known method for distributed version control in a domain-agnostic way. 
In this regard, we have chosen \emph{Git} for the following two reasons.
On the one hand, Git is a mature version control system supported by sophisticated tools and broadly used in software development projects.
More than 10 million repositories\furl{https://github.com/blog/1724-10-million-repositories} are hosted on GitHub for open source and commercial projects~\cite{kalliamvakou2015open}.
On the other hand, existing popular vocabularies like \emph{schema.org}\furl{{https://github.com/schemaorg/schemaorg}}, \emph{Description of a Project} (DOAP)\furl{{https://github.com/edumbill/doap}}, the \emph{music ontology}\furl{{https://github.com/motools/musicontology}} publish their efforts in GitHub to leverage the contribution of the community. 
This indicates that the vocabulary development community is already familiar with Git.

The remainder of this paper is structured as follows: 
In \autoref{Requirements} we present a comprehensive list of requirements aggregated from the current state of the art and our ongoing work on \emph{MobiVoc}\furl{{https://github.com/vocol/mobivoc}} and \emph{SCORVoc}\furl{{http://purl.org/eis/vocab/scor}}. 
In \autoref{git4Voc} we present \emph{Git4Voc} which comprises guidelines how Git can be used for collaborative vocabulary development. 
With Git4Voc we propose to utilize Git's hooks mechanism to realize vocabular-specific features.
In \autoref{proofOfConcept} we demonstrate concrete example of hook implementations. 
We provide an overview about related work in \autoref{relatedWork}.
The conclusion and an outlook to future work are presented in \autoref{conclFutureWork}. 

\section{Requirements of Collaborative Vocabulary Development}
\label{Requirements}

Collaborative vocabulary development is considered to be very related to the broad field of software development.
In fact, most proposals for supporting the former are inspired by experiences in the latter.
However, a vocabulary is not totally equal to software code. 
The development of vocabularies raises challenges which are new and not or at least not to that extend raised during software development.
In this section we focus on requirements which are more critical for vocabularies.
We gathered these requirements by aggregating insights from the current state of the art and our own experiences during the development of MobiVoc\furl{{https://github.com/vocol/mobivoc}} and SCORVoc\furl{{http://purl.org/eis/vocab/scor}}.
In the following these requirements are presented in detail.

\textbf{Communication support (R1)}
Collaborative vocabulary development is about finding consensus between members of a team.
In order to share ideas and finding agreements, communication among the contributors is essential~\cite{noy2006framework}.
During the whole life cycle, especially in agile development, supporting and recording discussions, changes and their reasons are crucial~\cite{noy2008collaborative}.
This is especially very important in the case of heterogeneous teams with experts from different domains. 
Some critical examples to be communicated within a team are introducing new elements, extending or modifying the subsumption hierarchy, integration of external resources and changing the underlying semantic expressivity~\cite{seidenberg2007state}.
An effective communication has a significant impact on the quality of the collaboration and its outcome.

\textbf{Provenance of information (R2)}
In collaborative development the capability to track the changes made by contributors is an important feature~\cite{noy2008collaborative}.
This is due to the fact that each change in the vocabulary reflects the understanding of the authors regarding the domain.
In case of disagreements, it is necessary to know which change was made by whom at which time and for what reason.

\textbf{Different roles (R3)}
Creating vocabularies with the purpose of realizing data integration across heterogeneous independent systems, involves domain experts from various fields with different levels of expertise.
For instance, in large projects like the \emph{Gene Ontology}\furl{{http://www.geneontology.org/}} (GO) many participants and curators take part in the development process. 
Most participants can only add comments and discuss terms. 
A core team is allowed to edit the main components of the vocabulary by adding modules, classes, properties, removing terms and performing refactoring. 
For that reason, there is a need for the definition of roles along with the permissions~\cite{seidenberg2007state, noy2008collaborative, tudorache2011knowledge, simperl2014collaborative}.

\textbf{Workflow independence (R4)}
The overall field of methodologies and workflows for collaborative vocabulary development is changing continuously~\cite{noy2008collaborative}.
To the best of our knowledge, there are no established methodologies nor workflows which are broadly applied.
Tools supporting collaboration should be generic and be able to adapt in highly dynamic context.
Therefore, it is important that a system is flexible enough to be used within different methodologies and workflows.

\textbf{Quality assurance (R5)}
Developing vocabularies includes many requirements of quality assurance.
Syntax and semantic correctness as well as the application of best practices on designing vocabularies are some of the quality aspects. 
Therefore providing tool support is a significant feature to prevent contributors from making errors. 
Later correcting phases might lead to a wasting of resources in terms of time and money.

\textbf{Documentation generation (R6)} 
As mentioned before, a team for vocabulary development comprises domain experts with less technical expertise in knowledge representation and engineering tools.
In order to enable them contributing to the development process, providing user friendly view to the current state of the vocabulary is vital. 
Therefore, an automatic documentation generation feature is necessary.

\textbf{Deltas among versions (R7)}
Collaborative development of vocabularies should respond to the evolution of the knowledge domain~\cite{simperl2014collaborative}.
It should also respect the evolution of connected vocabularies within the Linked Data Cloud, in order to avoid semantic inconsistencies.
Therefore, support for detecting and documenting the semantic difference between versions is needed, to enable developers to understand the mentioned evolutions.
This includes the modification, the addition of new elements (i.e. classes, properties) as well as the removal of existing terms.
Authors of well-known vocabularies such as \emph{SKOS}\furl{{http://www.w3.org/2004/02/skos/history}} and \emph{schema.org}\furl{{http://schema.org/docs/releases.html}} publish release notes containing what has been changed among different versions.

\textbf{Editor agnostic (R8)} 
In contrast to software code, vocabularies are abstract artefacts which can be serialized with different techniques.
Since contributors can use different editors which style the syntax in different ways, the support of the collaboration must be editor agnostic and syntax independent.

\textbf{Modularity (R9)}
Modularization is recognized as an important step in collaborative vocabulary building~\cite{suarez2012neon}.
Reusability, the decrease of complexity, ownership and customization are some of the benefits of vocabulary modularization.
Some studies report that there is no universal way to perform this process and that the choice of a particular technique should be guided by application specific requirements~\cite{d2007ontology}.
In contrast, other reports show that a module in a mid-sized vocabulary should contain between 200 and 300 lines of code~\cite{schlicht2006h}. 
Especially in an agile development process with large vocabularies and many contributors, it is of paramount importance that the system provides means to support the modularization activity.

\textbf{Multilinguality (R10)} 
In order to have a wide range of applicability to different cultures and communities, vocabulary terms \textit{must} be translated into various languages~\cite{gracia2012challenges}. 
The localization (and internationalization) process of vocabularies should be supported by the system. 

\textbf{Labeling versions (R11)} 
Release versions of vocabularies should be labeled appropriately. 
This ensures that users that can be humans or machines have always the possibility to use specific version, not only the latest one.


\section{Git4Voc}
\label{git4Voc}

In this section we present Git4Voc.
On the one hand, we propose guidelines how Git can be used for collaborative vocabulary development project.
On the other hand, we present how the requirements from \autoref{Requirements} can be technically implemented by Git \emph{hooks}.
Additionally, in terms of guidelines we analyzed best practices from collaborative software development and identified the following aspects as critical for the quality of the vocabulary: 
(1) management of generated information; 
(2) rights management; 
(3) branching and merging; 
(4) automate development and deployment tasks by hooks;  
(5) tool independence;
(6) vocabulary organization structure; and
(7) labeling of release versions. 
In the next subsections we show in detail how our approach responds to the above mentioned requirements.

\subsection{Management of Generated Information} 
\label{ManagementGeneratedInformation}

During the development process a bunch of information is generated by the contributors. 
The capability to manage this information within the entire project life-cycle is essential. 
In fact, value added services like GitHub, GitLab or BitBucket enrich Git functionality with powerful information management features.
For instance, issues are a great way of tracking communications, reporting problems as well as bug fixes and announcing version releases. 
Communities like \emph{schema.org} manage their discussions using GitHub.
The above mentioned means support requirement \textbf{(R1)}.
Based on this fact, we propose that activities gathered in \autoref{tabCommonVocOperations} should be documented.
If possible, the name of issues should correspond to the name of the activities.

Another important requirement in collaborative vocabulary development is the ability to view the history of the changes (called traceability in software engineering).
This addresses the requirement \textbf{(R2)}. 
Using commands \emph{git log} and \textit{git diff} a user can explore the history of the commits and the differences between them.
Each commit should be realized based on \emph{Best Commit Practices\furl{{http://www.git-tower.com/learn/git/ebook/command-line/appendix/best-practices}}}. 
In vocabulary development the \emph{atomicity} of commits is of paramount importance.

\subsection{Rights Management} 
\label{RightsManagement}

Standalone solutions such as \emph{GitLab}\furl{{https://gitlab.com/gitlab-org/gitlab-ce/blob/master/doc/permissions/permissions.md}} and \emph{Gitolite}\furl{{https://github.com/sitaramc/gitolite}} as well as third-party services like \emph{Bitbucket}\furl{{ https://confluence.atlassian.com/display/BITBUCKET/Add+Users,+Set+Permissions,+and+Review+Account+Plans}} and \emph{GitHub}\furl{{ https://help.github.com/articles/permission-levels-for-an-organization-repository}} offer basic options for user rights managements, like reading, writing, posting, adding new team members and adding tags. 
However, even with these solutions a high level of user management i.e. restricting editing a specified number or type of classes, properties or instances cannot be achieved with Git.
In order to address requirement \textbf{(R3)}, we explore a combination of branching and hooks.

\ctable[
  label={tab:roles},
	pos   = tb,
  caption={Different roles and their primary activities.},
]{lccc}{
}{\FL
  \textbf{Roles}		                 & \textbf{Basic} & \textbf{Semantic} & \textbf{Structural}	    \\
                		                 & \textbf{Activities} & \textbf{Issues} & \textbf{Issues}	    \ML
  Vocabulary Eng.      	 & +		       		& +            			& +     	   			\NN
  Domain Expert        		 & +            		& -            			& -		   			 \NN
  Users			     	     & -            		& -            			& -		   		 \NN
  Translators				 & +   		    		& - 			   			& -                        
  \LL}

With the combination of branching and hooks with role definition for users, fine grained access management can be achieved. 
Concretely, by using server-side hooks, realizing rights managements on top of user roles is possible.
For instance, an implementation of a \emph{pre-push} hook can check for the user's role and permissions and deny if the necessary rights according the activity and branch are not set. 

\autoref{tab:roles} shows common roles and their permissions, with respect to the defined categories of activities.
In a trusted environment right management can also be realized with client-side hooks. 
An example for this is depicted in \autoref{lst:RapperValidation}, where the user is denied to push to the master branch.

 \subsection{Branching and Merging} 
\label{branchingStrategy}

  \begin{table*}[tb]
  \caption{Common Activities in Collaborative Vocabulary Engineering}
  \label{tabCommonVocOperations}
	\centering
	\begin{tabular}{p{0.038\linewidth}p{0.12\linewidth}p{0.355\linewidth}p{0.4\linewidth}}
	\hline
  \textbf{Activity} 	    & \textbf{Name} 	    &\textbf{Description} 			              & \textbf{Example} \\
	\hline
	
  \textbf{ACT1}	  &
  Simple Addition/Deletion	  &
	Adding new or deleting existing elements like classes and properties   		      & 
	Adding a class in the last level of the taxonomy						 \\
	\hline
	
	\textbf{ACT2}	    &
  Complex Addition/Deletion	&
	Adding new elements to be interconnected within the existing class or properties taxonomy 		      & 
	Adding a object property as a super property of two existing properties 					 \\
	\hline
	
	\textbf{ACT3}	  &
  Modification		&
	Modifying existing elements 		&
	Modifying the domain and range of an existing object property \\ 
	\hline
	
	\textbf{ACT4}	  &
	Reusing		      &
	Reusing elements of the Linked Data Cloud 	&
	Defining new local concepts by using external resources\\
	\hline
	
	\textbf{ACT5}	  &
	Alignment		    &
	Alignment of existing elements with equivalents in the cloud 	&
	Alignment of classes and instances with \emph{owl:equivalentClass} and \emph{owl:sameAs}					\\
	\hline
	
	\textbf{ACT6}	  &
  Refactoring		  &
	Changing the name and metadata of an specific element and its connections	 		&
	Renaming a class which is connected in many domain and range relation of properties and need to be renamed everywhere\\
	\hline
	
	\textbf{ACT7}	  &
	Common Metadata		  &
	Adding/Removing/Modifying predefine \emph{RDFS} metadata  	 			     &
	Adding metadata to a class with \emph{rdfs:label, rdfs:comment} 				\\
	\hline
	
	\textbf{ACT8}	  &
  External Metadata		&
	Adding/Removing/Modifying external metadata   	 			       &
	Adding metadata to a class with \emph{skos:prefLabel, dc:title} 				\\
	\hline
	
	\textbf{ACT9}	  &
  Translating		   &
	Adding/Removing/Modifying translation for the terms 	    &
	Using \emph{rdfs:label} to translate elements into different languages \\
	\hline
	
	\textbf{ACT10}	  &
  Modularization   &
	Adding new modules to the existing vocabulary	 	 &
	Creating and integrating new modules due to new requirements \\
	\hline
	
	\textbf{ACT11}	  &
  Partitioning	   &
	Partitioning into different modules with existing elements	 			      &
  The vocabulary has grown in size and semantic complexity. 
	Partitioning the existing vocabulary into different modules		\\
	
	\hline
	\end{tabular}
	\end{table*}

Git is a very flexible tool, which addresses requirement \textbf{(R4)}. 
Using Git, teams are able to organize their work in different types of workflows\furl{{https://www.atlassian.com/git/tutorials/comparing-workflows/forking-workflow}}.
Branching strategies affect the quality in collaborative software development~\cite{phillips2011branching, shihab2012effect}. 
Vocabulary development is mostly accepted to be a specific type of software development.
Therefore, it is considered that the branching strategy affects the quality of the vocabularies. 
Well-known projects such as \emph{schema.org} use branches to organize their work.
In order to design a branching model, it is important to understand the possible activities that a team can perform. 
In this regard, we collected common activities of collaborative vocabulary development which are listed in \autoref{tabCommonVocOperations}.
Aiming at producing a vocabulary with good quality, the entire team should be aware of these activities and how to face them in the development process.
Due to their impact on the overall vocabulary, we have classified these activities into three categories: 
(1) basic activities (\emph{ACT1, ACT7, ACT9}), 
(2) semantic issues (\emph{ACT2, ACT3, ACT4, ACT5, ACT6, ACT8}) and 
(3) structural issues (\emph{ACT10, ACT11}).

\begin{figure}[tb]
  \centering    
  \includegraphics[width=1.05\linewidth]{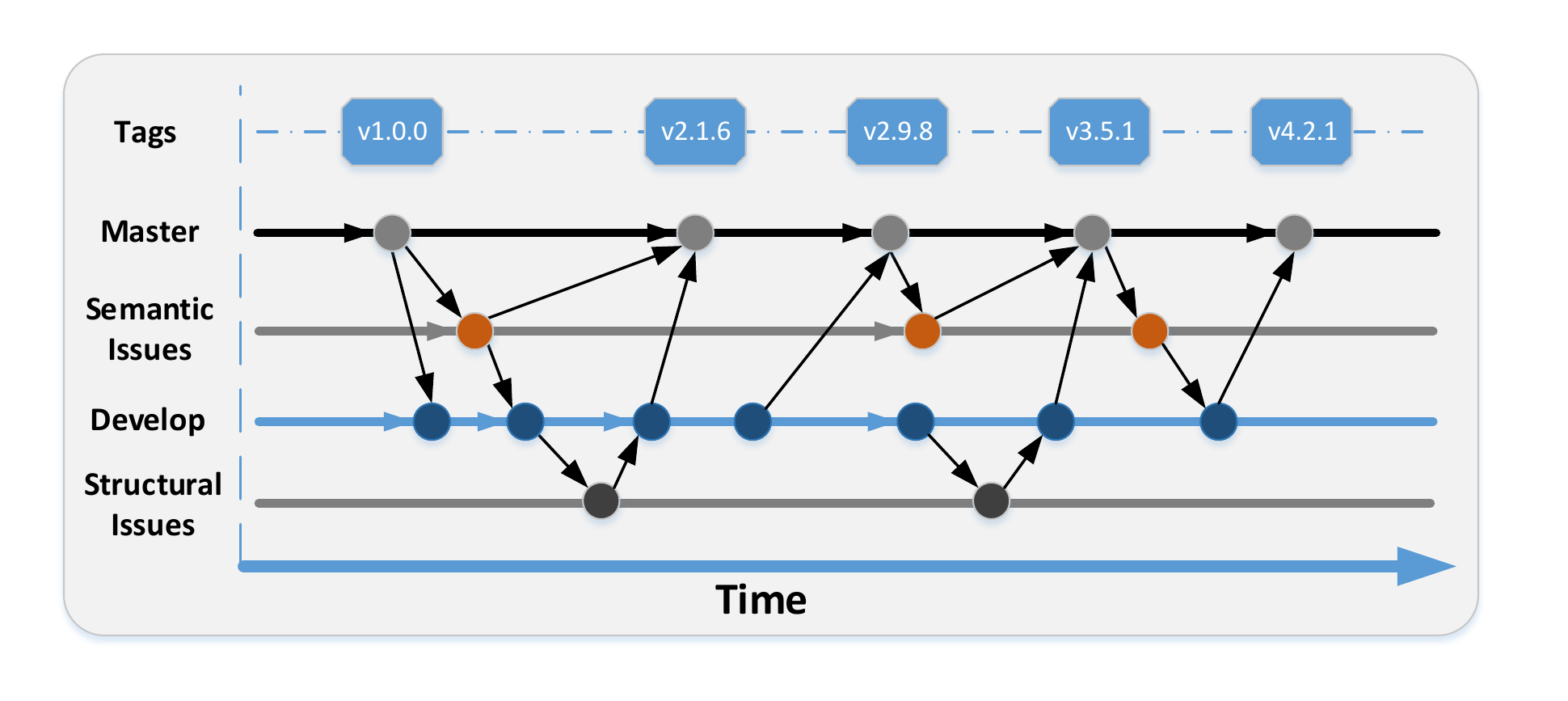}
   \caption{Branching model for vocabulary development}
    \label{fig:branching}
\end{figure}

This led us to the branching model that is depicted in \autoref{fig:branching}.
We designed different branches to handle the mentioned categories.
Basic activities have to be performed in the \textit{Develop Branch}. 
For the second category we propose a dedicated branch called \textit{Semantic Issues}.
In case of the third category a branch named \textit{Structural Issues} has to be applied. 
It is important to bear in mind that we are not restricting the flexibility of Git regarding branches.
On the contrary, other branches can be used as a complement of this model.
Nevertheless, our approach of branching model will help developers because those branches are connected to specific activities in collaborative vocabulary development. 
 
Our solution is built on top of the best practices for branching in software development\furl{{http://nvie.com/posts/a-successful-git-branching-model}}. 

\subsection{Automate Development and Deployment Tasks by Hooks}
\label{hooks}

\begin{figure}[tb]
  \centering    
  \includegraphics[width=0.8\columnwidth]{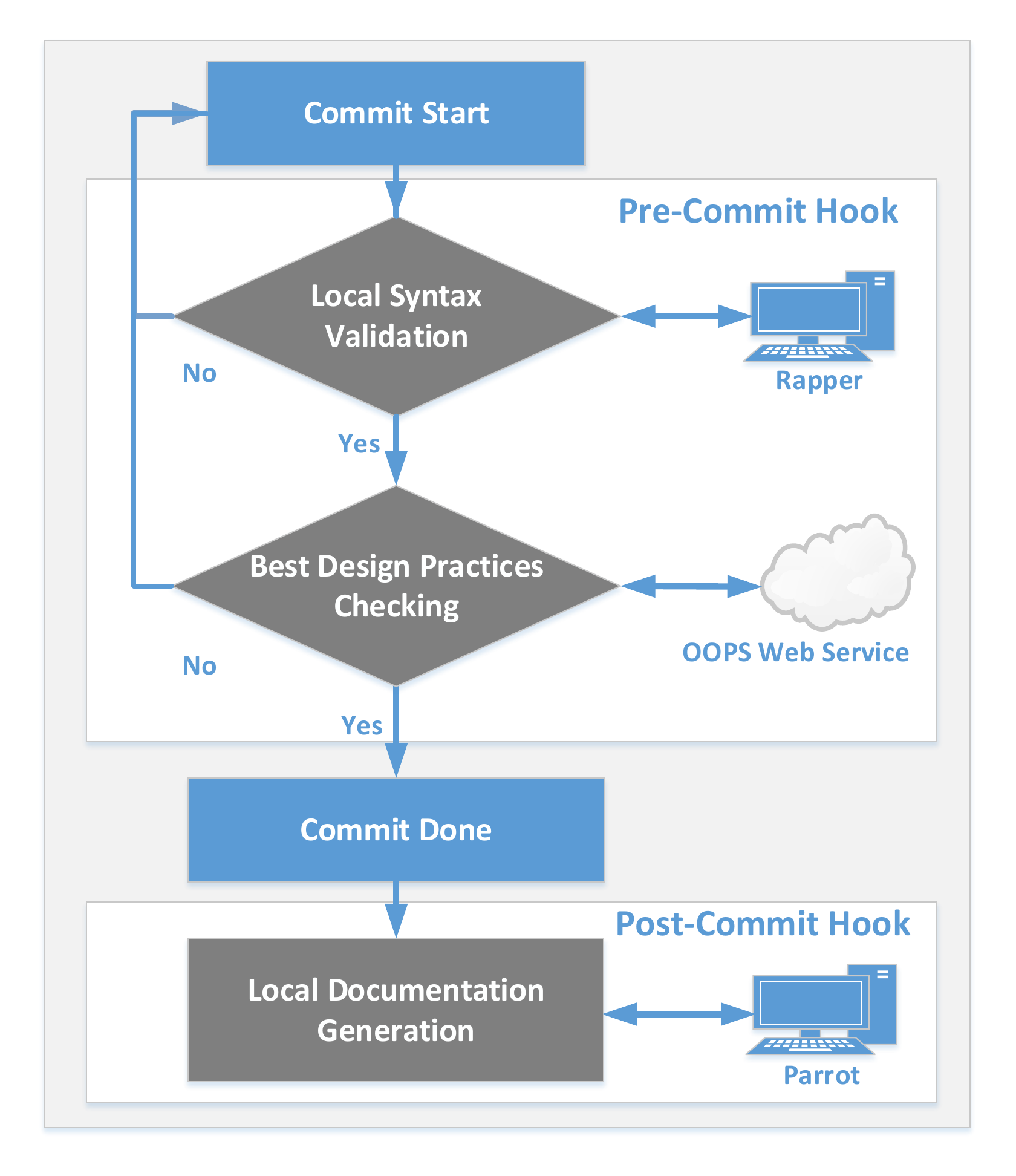}
   \caption{Client-side \textit{Hooks} Workflow}
    \label{fig:workflow}
\end{figure}

Despite the fact that Git has many implemented features, it allows extending its functionality by using so-called hooks. 
This is a mechanism that allows running scripts before or after specific Git events. 
Based on the execution place, two types of hooks are distinguished, client-side and server-side hooks. 
Due to space limitations, the examples in this article showcase only client-side hooks.
In order to address requirements \textbf{(R5)} and \textbf{(R6)}, we implemented the following three important tasks for collaborative vocabulary development: 
(1) syntax checking, 
(2) assessing vocabularies against best design practices and 
(3) documentation generation.
\autoref{fig:workflow} illustrates how these examples are integrated into the \emph{commit} process.


After modifying the local vocabulary and adding changes to the stage phase, the next step is to \emph{commit} the current state to the local repository. 
The initialization of commit triggers a hook named \textit{pre-commit}.  
\autoref{lst:RapperValidation} shows our implementation of this hook which realizes the tasks syntax checking and best practice assessment.
First, it retrieves all modified files with extensions such as \textit{rdf}, \textit{owl}, \textit{ttl} and checks for syntax errors by using \textit{Rapper}\furl{{http://librdf.org/raptor/}}. 
In case that vocabularies fail to pass the validation process, the commit is canceled. 
The user is notified with a message which shows detailed description about the error which comprised of the file name, line number and the error type.
If syntax validation is passed successfully, the modified files are posted to the \emph{OOPS}\furl{{http://oops-ws.oeg-upm.net/}} Web Service through
\textit{curl}, a command-line HTTP client. 
This service assesses vocabulary files for certain quality metrics. 
The result of this is a descriptive message that contains recommendations of best practices for vocabulary development.
If no errors exist, the pre-commit hook is finished and the commit is accepted. 
Afterwards a \textit{post-commit hook} is called. 
\autoref{lst:PostCommitHook} demonstrates our implementation of a post-commit hook for documentation generation in a human friendly format. 
This script uses \emph{Parrot}\furl{{https://bitbucket.org/fundacionctic/parrot/}} as an external tool.

For security reasons Git repository services do not allow to automatically distribute predefined \textit{hooks} on cloning phase. 
In order to accomplish this task, the repository itself should have a dedicated folder that contains the implemented hooks.
After the first clone, these hooks need to be copied to the \textit{.git/hooks} directory.
For that purpose, we implemented a script which needs to be executed after cloning the repository. 
Once this process is finished, predefined \textit{hooks} will be automatically executed after each commit.
However, when the hooks have been changed, e.g. to use different validation or documentation generation tools, this script has to be executed again.
Apart from installing the hooks, this script can also be used to download and install tools like Rapper, which are necessary for the hooks.  
If these tools are to be placed within the local repository, the file \textit{.gitignore} should be used to prevent them from being pushed to the remote repository.

Git does not show semantic diffs between versions of vocabulary.
\textit{Owl2VCS}~\cite{gonccalves2012ecco} shows deltas among different versions.
By using such a tool and hooks, generated deltas can be published is human friendly format as well.
This corresponds to the requirement \textbf{(R7)}.

\subsection{Tool Independence}
Collaborative working with Git can be facilitated by using vocabulary editors like \emph{Prot\'eg\'e}\furl{{http://protege.stanford.edu/}}, \emph{TopBraid Composer}\furl{{http://www.topquadrant.com/tools/IDE-topbraid-composer-maestro-edition/}}, \emph{Neon Toolkit}\furl{{http://neon-toolkit.org/wiki/Main_Page.html}}. 
As each of them has different algorithms for writing files, there might arise consistency problems in case that contributors are not using the same editor. 
For instance, one contributor use Prot\'eg\'e, whereas another one uses Neon Toolkit.
They are editing the same file simultaneously. 
After saving it, different representations of that file will be created. 
As a consequence Git recognizes lot of changes and asks for conflict resolution. 
This is due to fact that Git is a version control based on text line changing.
It detects when a line has been changed from the previous version.
In such a case using the merge tool is necessary, which is a time consuming and error prone task that could lead to information lose.

In order to avoid the above mentioned problems, we propose the use of Turtle format.
This addresses the requirement \textbf{(R8)}.
A similar approach describes a pattern to express data on GitHub storing it in CSV files\furl{{http://blog.okfn.org/2013/07/02/git-and-github-for-data/}}.
\autoref{lst:Turtle} presents our proposal to write one triple per line.

\begin{lstlisting}[style=turtle, caption={One triple per line}, label={lst:Turtle}]
@prefix rdf:  <http://www.w3.org/1999/02/22-rdf-syntax-ns#>.
@prefix rdfs: <http://www.w3.org/2000/01/rdf-schema#>.
@prefix owl:  <http://www.w3.org/2002/07/owl#>.
@prefix scor: <http://purl.org/eis/vocab/scor#>. 
@prefix vs:   <http://www.w3.org/2003/06/sw-vocab-status/ns#>.

 scor:Process
    rdf:type owl:Class ;
    rdfs:comment "A process is a unique activity..."@en ;
    rdfs:label "Process"@en ;
    rdfs:isDefinedBy scor: ;
    vs:term_status "testing".
      
 scor:Enable 
    rdfs:subClassOf   scor:Process ;
    rdfs:comment            "Enable describes the ...";
    rdfs:label              "Enable".    
\end{lstlisting}

\subsection{Vocabulary Organization Structure} 
\label{VOStructure}

Git's basic functionalities do not support modularizing code or vocabularies. 
Therefore, in order to address the requirement \textbf{(R9)}, we propose some guidelines for organizing the vocabulary in files where each file represents a module.
Considering the fact that each line should represent a triple and based on the insights on ~\cite{schlicht2006h}, we propose that files should not contain more than 300 triples.
We highlight three possible forms of organizing the files.
All of these cases use single Git repository to store the files. 

\emph{1. The complete vocabulary is contained in one single file.} 
When the vocabulary is small (e.g. contains less than 300 lines of code) and represents a domain which cannot be divided in sub domains, it should be saved within one single file.
If the number of contributors is relatively small and the domain of the vocabulary is very focused, organizing it into one single file might be possible, even if it exceeds 300 lines of code.
However, if the comprehensibility is exacerbated, splitting it into different files should be considered.

\begin{figure}[tb]
  \centering    
  \includegraphics[width=\linewidth]{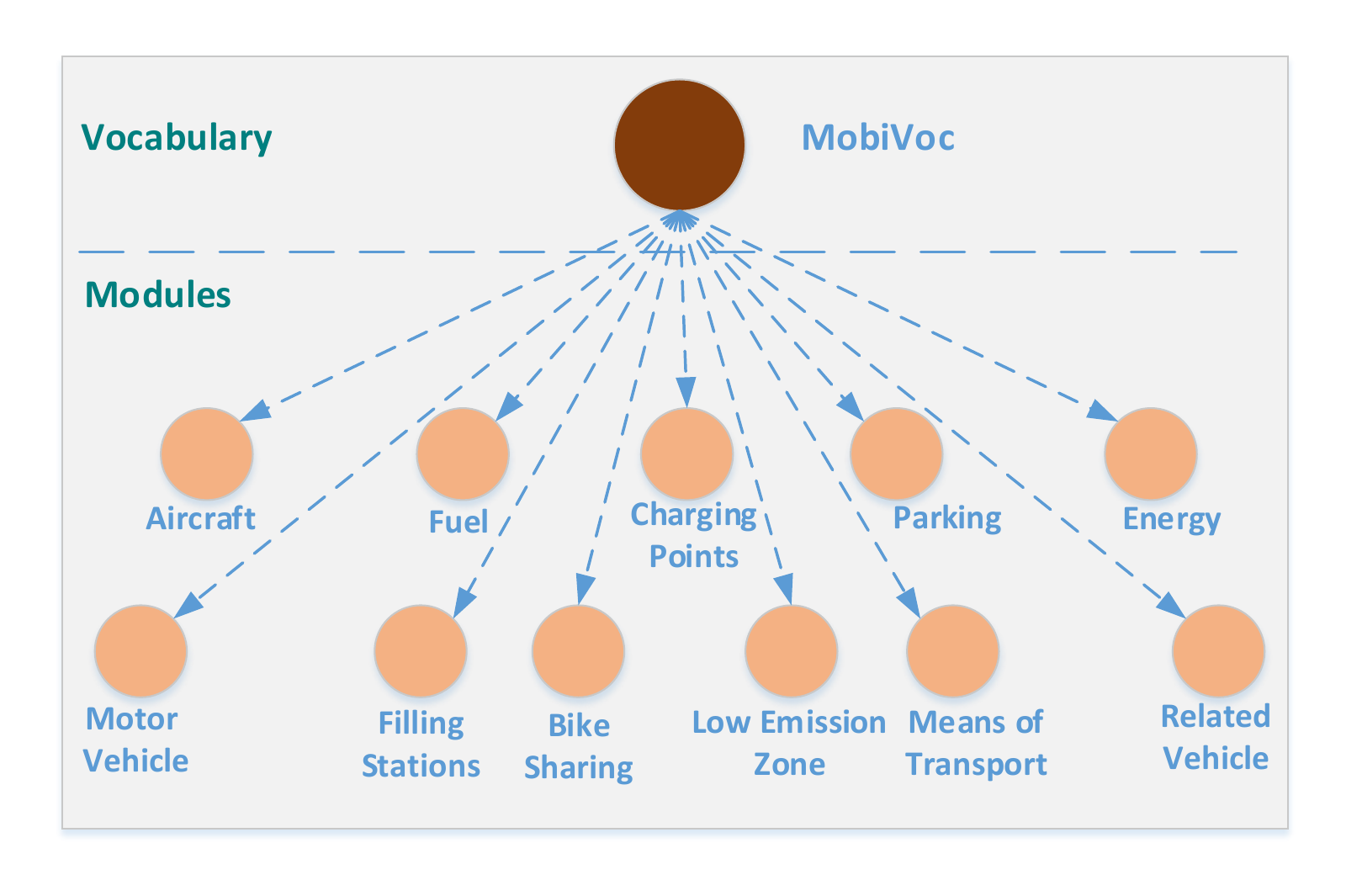}
   \caption{The structure of MobiVoc}
    \label{fig:MobiVocModules}
\end{figure}

\emph{2. The vocabulary is split in multiple files.}
If the vocabulary contains more than 300 lines of code or covers a complex domain, it should be organized into different sub domains or modules.
In this regard, we mapped sub domains with modules.
When the sub domains themselves are small enough they should be represented by different files within the parent folder.
There exists patterns for vocabulary modularization~\cite{abbes2012characterizing}.
We developed the MobiVoc based on the pattern \textit{n modules importing 1 module}.
In this case, \textit{1 module} was the vocabulary itself.
The \textit{n modules} like Aircraft, Fuel were saved in separate files.
Each file represents a specific sub domain.
By following this approach, domain experts can contribute independently to vocabulary development according to their field of expertise.
\autoref{fig:MobiVocModules} depicts the structure of MobiVoc and its modules.

\emph{3. Vocabulary modules are stored in files and folders. }
For huge vocabularies that comprises complex domains, splitting it into files is not sufficient.
This would lead to a large amount of files within a single folder.
Therefore, if the sub domains are large enough to be split into files they should be represented by folders.
Each folder contains files which represents modules.
In this case, the folder and file structure should reflect the complex hierarchy of the overall domain. 

Through splitting the vocabulary in files for specific purposes, the requirement \textbf{(R10)} is addressed as well.
This can be achieved by creating dedicated files for translating.
In these files users with the role \textit{Translators} can contribute by translating the terms into the required language.

\subsection{Labeling of Release Versions} 
\label{VersionLabelling}

Based on requirement \textbf{(R11)}, proper labeling of release versions is vital, as it facilitates re-usability. 
One of the common ways to realize that is to deploy each release version in different files. 
However, this could lead to following problems as identified in~\cite{TemporalTags}: 
(1) the number of files could increase rapidly, 
(2) choosing versions creates confusion, 
(3) maintenance needs additional resources and 
(4) synchronizing with latest version from dependent applications requires additional effort.
To avoid the above mentioned problems, we have kept versions of vocabularies in the same file.
These versions are separated by Git implemented functionality of tagging and saved in the master branch which is part of the branching model and illustrated in \autoref{fig:branching}.
It is possible to create and filter tags at any time. 
Moreover, users can obtain a specific version of the vocabulary just by giving the tag name. 
Therefore, each released version of a vocabulary \emph{must} have a version number. 
Based on the scheme from~\cite{VersVocabLDW} and the mentioned categories of activities in \autoref{tabCommonVocOperations}, we propose tagging different versions according to the following pattern: 
\emph{v[StI.SeI.BA]}, where \emph{StI} stands for Structural Issues, 
\emph{SeI} for Semantic Issues and 
\emph{BaA} for Basic Activities.
Each category is related with a number, in the respective position.
Changes in the vocabulary regarding to the categories are commonly reflected by increasing the numbers. 
For instance, the difference between releases \emph{v[1.0.0]} and \emph{v[2.0.0]} shows structural issue changes (\emph{StI}).

\section{Implementation}
\label{proofOfConcept}

We have developed Git4Voc\footnote{\url{https://github.com/vocol/vocol/tree/master/Git4Voc}}, which is an environment for collaborative vocabulary development.
~\autoref{tab:requirements} provides an overview which of the previously described requirements are fulfilled by Git4Voc.
This solution combines Git4Voc with a set of state-of-the-art tools like Rapper, OOPS Service and Parrot.
Each tool is exchangeable and can be easily replaced by alternatives.
They provide services which are called by the hooks mechanism.
In the following these hooks are presented in detail.

\begin{table*}[tb]
  \caption{Collaborative Requirements from Git4Voc perspective.}
  \label{tab:requirements}
	\centering
	\begin{tabular}{p{0.02\linewidth}p{0.19\linewidth}>{\centering\arraybackslash}p{0.09\linewidth}p{0.6\linewidth}}
	\hline
  \textbf{No.} 	&\textbf{Requirement} 			& \textbf{Supported} 		 		& \textbf{Means} 	    \\
  				& 								& \textbf{by plain Git}						& \\
	\hline
  R1		&Communication support 		& +						 					& Issues tracking system offered by hosting platforms GitHub, GitLab, BitBucket		\\
	
  R2		&Provenance of information 		& + 						 					& Git log and git diff functionality		\\
	
  R3		&Different roles	 			     & + 				 						& Further extension by a combination of branching and hooks	\\
	
  R4 	&Workflow independence 				& + 			 		 						& Git, branching and merging strategies		\\
	
  R5 	&Quality assurance 				&  - 				 						& Extended by a combination of hooks and tools like: Rapper, JenaRiot		\\
	
  R6		&Documentation generation	 		&  - 		 							&  Extended by a combination of hooks, documentation generation tools like: SchemaOrg, Widoco \\
  
  R7		&Deltas among releases	 		&  - 		   									& Extended by a combination of Owl2VCS as an external tool and hooks\\

  R8	&Editor agnostic	 				&  + 		 							&   Git base functionality\\
  	
  R9 	&Modularity			 		&  + 		 	 						&  Hierarchy organization of vocabularies where modules are represented by files\\
		
  R10		&Multilinguality	 				&  + 		 							&   Hierarchy organization of vocabularies where dedicated files are used for multilinguality\\
  
  R11	&Labeling versions				&  + 		  						&   Git tag functionality\\
  
	\hline
	\end{tabular}
	\end{table*}

The \autoref{lst:InstallScript} shows an example how predefined hooks are copied into the \textit{.git/hooks} folder after cloning the repository.
In addition, it shows installing of the tools: Raptor and Parrot and their necessary libraries in case they do not exist. 

\begin{lstlisting}[language=bash,breaklines=true, caption={Install Hooks and Tools}, ,label={lst:InstallScript}]
#!/bin/sh
# Copy the modified hooks 
for i in `ls -1 hooks` do
	cp hooks/$i .git/hooks/$i
done
# Create directory for necessary tools
if [ -d "$tools" ]; then
  mkdir -p "tools"
  ...
fi
cd tools

#Install Raptor
if [ ! -d "$Raptor" ]; then
  curl -O http://download.librdf.org/source/raptor2-2.0.15.tar.gz
  ...
  sudo apt-get install libxml2-dev libxslt1-dev python-dev
  sudo apt-get -y install raptor2-utils
  ...
fi

#Install Parrot
if [ ! -e "parrot-jar-with-dependencies.jar" ]; then
  curl -O https://github.com/vocol/vocol/raw/master/Hooks/tools/parrot-jar-with-dependencies.jar
fi

...

\end{lstlisting}

The \textit{pre-commit hook} is adapted to prevent users from committing to the master branch as shown in the \autoref{lst:RapperValidation}. 
This example can be further customized to restrict committing to other branches as well.
By doing so, a low level of rights management is achieved on the local repository, before the changes are pushed to the remote repository.
Furthermore, to reduce the efforts needed for subsequent corrections, we integrated tools for (1) syntax validation; and (2) checking for bad modeling practices. 
For the first, the Rapper tool is used, which validates each turtle file for syntactic errors. 
For the second, we used OOPS Web Service to scan vocabulary files for bad modeling practices.

\begin{lstlisting}[language=bash,breaklines=true, caption={Pre-Commit Hook: Syntax validation and checking for bad modeling practices} ,label={lst:RapperValidation}]
#!/bin/bash
currentBranch=$(git symbolic-ref HEAD)

if [ "$currentBranch" = "refs/heads/master" ]; then
	 echo "Not allowed to commit to master branch!"
	 exit 1
else
	# Get only modified files with ttl extension
	files=$(git diff --cached --name-only --diff-filter=ACM | grep ".ttl$")
	...
	
	for file in ${files}; do
        # Validate each file using Rapper
	    res=$(rapper -i turtle "${files}" -c 2>&1)
    ...
    
    if ! $error; then
          for file in ${files}; do
           fileContent=`cat ${files}`
	  	   request="<?xml ...>"
	       res=$(curl -X POST -d "$request" -H "Content-Type: application/xml" http://oops-ws.oeg-upm.net/rest)
	       ...
	  done
	fi

	if ${succeed}; then
	    echo "COMMIT SUCCEEDED"
	else
	    echo "COMMIT FAILED"
	    exit 1
	fi
fi
\end{lstlisting}

The \autoref{lst:PostCommitHook} shows the \textit{post-commit hook}, which uses Parrot to create the human friendly representation of the developed vocabulary.
The generated content is saved as a single HTML file which consists of all vocabulary elements.
User is able to navigate through entire vocabulary by merely selecting the element name.
Moreover, in order to create different representation style of the vocabulary, alternative tools such as: \emph{Widoco}\furl{{https://github.com/dgarijo/Widoco}}, \emph{Specgen}\furl{{https://github.com/specgen/specgen}}, \emph{Dowl}\furl{{https://github.com/ldodds/dowl}}, etc. (c.f. commented part of the code) can be used as well.

\begin{lstlisting}[language=bash,breaklines=true, caption={Post-Commit Hook: Documentation Generation}, ,label={lst:PostCommitHook}]
#!/bin/sh
files=$(git diff --cached --name-only --diff-filter=ACM | grep ".ttl$")
if [ "$files" = "" ]; then 
    exit 0 
fi	

for file in ${files}; do
    # Generate documentation using Parrot
    java -jar @path/parrot-jar-with-dependencies.jar -i "${file}" -o "${file}".html
    # Generate documentation using Widoco
    # java -jar  @path/widoco-0.0.1-jar-with-dependencies.jar -ontFile "${file}" -outFolder /home/
    ...
done
echo "\Documentation Generation is completed.\n"
exit 1
\end{lstlisting}

\section{Related Work}
\label{relatedWork}

Collaborative vocabulary development is an active research area in the Semantic Web community~\cite{palma2011holistic}.
Existing approaches like \emph{WebProt\'eg\'e}~\cite{WebProtegeSWJ} provides a collaborative web frontend for a subset of the functionality of the Prot\'eg\'e OWL editor. 
The aim of WebProt\'eg\'e, is to lower the threshold for collaborative ontology development.
\emph{Neologism}~\cite{basca2008neologism} is a vocabulary publishing platform, with a focus on ease of use and compatibility with Linked Data principles.
Neologism focuses more on vocabulary publishing and less on collaboration.
VocBench~\cite{stellato2015vocbench}, is an open source web application for editing thesauri complying with the SKOS and SKOS-XL standards. 
VocBench has a focus on collaboration, supported by workflow management for content validation and publication. 


The main limitation of the aforementioned tools is the lack of version control.
Therefore, we only consider approaches focused on using version control systems for collaborative vocabulary development. 

\emph{SVoNt}~\cite{luczak2010svont} extends the functionality of Apache SubVersion (SVN) by providing a possibility for versioning OWL conform lightweight description logic. 
SVN manipulates only with deltas of files, therefore SVoNt use a separate server to create conceptual changes between versions of ontologies. 
These changes are generated as a result of \textit{diff} operation between the modified ontology and the base ontology.
\emph{ContentCVS}~\cite{jimenez2009contentcvs} is a Prot\'eg\'e plugin. 
It adapts concepts from \emph{concurrent versioning} to enable developers to work in parallel.  
Moreover, it has features for conflict detection and resolution by checking structure and semantic of the ontology versions.
In \cite{VersVocabLDW} is described how the developers of \emph{RDA Vocabularies}\furl{{http://www.rdaregistry.info}} adopt rules from \emph{SemVer}\furl{{http://www.semver.org}} to realize a meaningful versioning using Git. 
Additionally, it provides general notes for organizing the vocabulary development in branches. 
\cite{zaikin2013owl2vcs} describes Owl2VCS, a toolset designed to facilitate version control of OWL 2 ontologies using version control systems. 
It can be integrated as an external tool with Git, Mercurial and Subversion and provide algorithms for structural diff~\cite{gonccalves2012ecco}. 
However, none of the above mentioned approaches cover all the identified requirements (c.f. \autoref{Requirements}) for collaborative vocabulary development.
On the contrary, our work analyze and address each one of them by using Git and Git4Voc as an extension.

\section{Conclusion and Future Work}
\label{conclFutureWork}

In this paper, we investigated the applicability of Git for collaborative vocabulary development.
We defined collaborative vocabulary development as the process of identifying the main terms across heterogeneous data sources by finding a consensus between the developers. 
The main challenge in this regard is the realization of a powerful collaborative environment. 
Distributed version control systems enable developers around the world to work collaboratively on complex software systems.
Since software and vocabularies are not the same, we analyzed their differences in detail by identifying requirements for a version control system that supports collaborative vocabulary development.
Our approach extends plain Git functionality by utilizing the hooks mechanism in combination with external tools to address these requirements.	
The presented approach is easily extensible and can accommodate additional external tools.

Regarding the future work, we are going to extend our approach with the full implementation of \textit{server side hooks}.
By doing so, tasks like: deploying specific versions of vocabularies to a dedicated server, generating deferencable URI's, ontology partitioning and modularization tasks can be performed in a fully automated way.
We also plan to develop and integrate a tool that validates vocabularies against conventions~\cite{grangel2015Convention} and provides recommendations for solving possible issues.
This will lead to a convenience and less error prone collaborative vocabulary development environment. 
As a result, all generated artefacts will be publicly accessible from all interested parts.

\section*{Acknowledgments}
This work is supported by the German Ministry for Education and Research funded project LUCID and European Commission under the Seventh Framework Program FP7 for grant 601043 (\url{http://diachron-fp7.eu}).

\bibliographystyle{IEEEtran}
\bibliography{BIBReferences}

\end{document}